\documentclass{article}

\usepackage[final]{corl_2019} 

\usepackage{graphicx}
\usepackage{subfigure}
\usepackage{bm}
\usepackage{amsfonts}
\usepackage{amsmath}
\usepackage{algorithm}
\usepackage{algpseudocode}
\usepackage{cancel}

\newtheorem{lemma}{Lemma}
\newtheorem{proposition}{Proposition}
\newtheorem{theorem}{Theorem}
\newtheorem{corollary}{Corollary}
\newtheorem{definition}{Definition}

\DeclareMathOperator*{\argmax}{arg\,max}
\DeclareMathOperator*{\argmin}{arg\,min}
\DeclareMathOperator*{\relu}{relu}

\title{Mutual-Information Regularization in Markov Decision Processes and Actor-Critic Learning}

\author{
  Felix Leibfried, \, Jordi Grau-Moya\\
  PROWLER.io\\
  Cambridge, UK\\
  \texttt{\{felix,jordi\}@prowler.io} \\
}

\begin{document}
\maketitle

\begin{abstract}
Cumulative entropy regularization introduces a regulatory signal to the reinforcement learning (RL) problem that encourages policies with high-entropy actions, which is equivalent to enforcing small deviations from a uniform reference marginal policy. This has been shown to improve exploration and robustness, and it tackles the value overestimation problem. It also leads to a significant performance increase in tabular and high-dimensional settings, as demonstrated via algorithms such as soft Q-learning (SQL) and soft actor-critic (SAC).
Cumulative entropy regularization has been extended to optimize over the reference marginal policy instead of keeping it fixed, yielding a regularization that minimizes the mutual information between states and actions. While this has been initially proposed for Markov Decision Processes (MDPs) in tabular settings, it was recently shown that a similar principle leads to significant improvements over vanilla SQL in RL for high-dimensional domains with \emph{discrete} actions and function approximators.

Here, we follow the motivation of mutual-information regularization from an inference perspective and theoretically analyze the corresponding Bellman operator. Inspired by this Bellman operator, we devise a novel mutual-information regularized actor-critic learning (MIRACLE) algorithm for \emph{continuous} action spaces that optimizes over the reference marginal policy. We empirically validate MIRACLE in the Mujoco robotics simulator, where we demonstrate that it can compete with contemporary RL methods. Most notably, it can improve over the model-free state-of-the-art SAC algorithm which implicitly assumes a fixed reference policy.
\end{abstract}


\section{Introduction}
\label{sec:intro}

In RL and MDPs, agents aim at collecting maximum reward yielding optimal policies that are \emph{deterministic}. One way of obtaining \emph{non-deterministic} optimal policies is by introducing a regulatory signal that penalizes deviations from a stochastic reference policy~\cite{Azar2011,Rubin2012,Fox2016,Neu2017,Schulman2017,Leibfried2018}. For a uniform reference policy, cumulative entropy regularization is recovered~\cite{Ziebart2010,Haarnoja2017,Nachum2017,Haarnoja2018,Levine2018,Haarnoja2019}. The effect of the latter on RL is: improved exploration and robustness, and an alleviation of overestimated values~\cite{Fox2016,Haarnoja2018,Leibfried2018}. Importantly, overall performance is improved not only in the tabular setting~\cite{Fox2016}, but also in high-dimensional settings with function approximators as demonstrated via algorithms such as SQL and SAC~\cite{Schulman2017,Haarnoja2018,Leibfried2018}.
In~\cite{Tishby2011,Grau-Moya2019}, it has been proposed to optimize over the reference policy rather than keeping it fixed, yielding a constrained reward maximization problem with a constraint on the mutual information between states and actions in each time step. This has been investigated in the tabular setting by~\cite{Tishby2011} for MDPs, and in high-dimensional domains and RL by~\cite{Grau-Moya2019} where it was shown that for discrete actions, mutual-information regularization can lead to significant improvements over a fixed reference policy (i.e. SQL~\cite{Schulman2017,Leibfried2018}). 
Here, we follow the RL-as-inference perspective where optimizing for the reference policy increases the log marginal likelihood. Our contributions are threefold. A) we theoretically analyze the mutual-information-regularized Bellman operator. B) we then develop a mutual-information regularized actor-critic learning (MIRACLE) algorithm for continuous action spaces based on this Bellman operator. And C) we demonstrate that MIRACLE can attain results competitive with contemporary methods in challenging large-scale robotics domains from Mujoco---e.g. improving over the state-of-the-art SAC~\cite{Haarnoja2018} in Ant.

\section{Background}
\label{sec:background}

In the following, we provide some background and notation for MDPs and RL, see Section~\ref{sec:MDP_and_RL}. Subsequently in Section~\ref{sec:rational_inattention}, we focus on mutual-information regularization in non-sequential decision-making scenarios paving the way for the sequential decision-making setting discussed in Section~\ref{sec:mdp}.

\subsection{Markov Decision Process (MDP) and Reinforcement Learning (RL)}
\label{sec:MDP_and_RL}
 An MDP is a five-tuple $(\mathcal{S}, \mathcal{A}, P, R, \gamma)$ where $\mathcal{S}$ is the state set and $\mathcal{A}$ the action set. $P: \mathcal{S} \times \mathcal{A} \times \mathcal{S} \rightarrow [0;1]$ refers to the probabilistic state-transition function that maps state-action pairs $(\bm{s},\bm{a}) \in \mathcal{S} \times \mathcal{A}$ to next states $\bm{s}^\prime \in \mathcal{S}$ according to the conditional probability distribution $P(\bm{s}^\prime|\bm{s},\bm{a})$. The reward function $R: \mathcal{S} \times \mathcal{A} \rightarrow \mathbb{R}$ determines the instantaneous reward $R(\bm{s},\bm{a})$ when taking action $\bm{a} \in \mathcal{A}$ in state $\bm{s} \in \mathcal{S}$. The hyperparameter $\gamma \in (0;1)$ is a discount factor for discounting future rewards. In the MDP setting, an agent is specified by a behavioral policy $\pi: \mathcal{S} \times \mathcal{A} \rightarrow [0;1]$ that maps states to actions probabilistically according to the conditional probability distribution $\pi(\bm{a}|\bm{s})$. A policy $\pi$ is evaluated based on its expected future cumulative reward captured by the state value function $V^\pi$ and the state-action value function $Q^\pi$ respectively as:
\begin{equation}
\label{eq:value_functions}
V^\pi(\bm{s}) := \mathbb{E}_{\pi,P} \left[ \sum_{t=0}^\infty \gamma^t R(\bm{s}_t, \bm{a}_t) \middle| \bm{s}_0 = \bm{s} \right] \; \text{ and } \; Q^\pi(\bm{s},\bm{a}) := R(\bm{s},\bm{a}) + \gamma \mathbb{E}_{P} \left[ V^\pi(\bm{s}^\prime) \right] ,
\end{equation}
 where $t$ is the time index and $\bm{s}^\prime$ refers to the next state. The goal is to identify \emph{optimal} policies $\pi^\star$ that maximize expected future cumulative reward, yielding optimal value functions $V^\star$ and $Q^\star$ as:
\begin{equation}
\label{eq:optimal_value_functions}
V^\star(\bm{s}) := \max_\pi \mathbb{E}_{\pi,P} \left[ \sum_{t=0}^\infty \gamma^t R(\bm{s}_t, \bm{a}_t) \middle| \bm{s}_0 = \bm{s} \right] \; \text{ and } \; Q^\star(\bm{s},\bm{a}) := R(\bm{s},\bm{a}) + \gamma \mathbb{E}_{P} \left[ V^\star(\bm{s}^\prime) \right] .
\end{equation}

In the RL setting, agents do not have prior knowledge about the transition function $P$ and the reward function $R$, and need to identify optimal policies via environment interactions. A popular class of RL algorithms that are suited for high-dimensional environments and continuous action spaces are actor-critic algorithms that learn a parametric policy $\pi_\phi$ with policy parameters $\phi$ as:
\begin{equation}
\label{eq:actor_critic}
\max_\phi \mathbb{E}_{p,\pi_\phi} \left[ Q^{\pi_\phi}(\bm{s},\bm{a}) \right] ,
\end{equation}
where $Q^{\pi_\phi}$ is the state-action value function of $\pi_\phi$ and $p: \mathcal{S} \rightarrow [0;1]$ is a distribution over states. Since $Q^{\pi_\phi}$ can usually not be computed in closed form, it is approximated via function approximators, policy rollouts or a combination of both. The state distribution $p$ refers either to an empirical distribution induced by a replay memory that collects visited states~\cite{Degris2012,Lillicrap2016,Abdolmaleki2018,Fujimoto2018,Haarnoja2018,Haarnoja2019} or to the stationary state distribution when executing $\pi_\phi$ in the environment~\cite{Sutton2000,Schulman2015,Schulman2017b}.

\subsection{Mutual-Information Regularization for Non-Sequential Decision-Making}
\label{sec:rational_inattention}
A non-sequential decision-making scenario can be described via a four-tuple $(\mathcal{S},\mathcal{A}, R, p)$ comprising a state set $\mathcal{S}$, an action set $\mathcal{A}$, a reward function $R:\mathcal{S} \times \mathcal{A} \rightarrow \mathbb{R}$ and a state distribution $p: \mathcal{S} \rightarrow [0;1]$ from which states are sampled according to $\bm{s} \sim p(\bm{s})$. A behavioral policy is specified the same way as in the sequential setting as $\pi: \mathcal{S} \times \mathcal{A} \rightarrow [0;1]$ yielding a conditional probability distribution $\pi(\bm{a}|\bm{s})$ that maps states $\bm{s} \in \mathcal{S}$ to actions $\bm{a} \in \mathcal{A}$. The mutual-information regularized decision-making problem is then defined in its constrained and unconstrained form respectively as:
\begin{equation}
\label{eq:rationa_inattention}
\max_\pi \mathbb{E}_{p,\pi} \left[ R(\bm{s},\bm{a}) \right] \; \text{s.t.} \; I(\bm{S},\bm{A}) \leq C \; \; \Longleftrightarrow \; \; \max_\pi \mathbb{E}_{p,\pi} \left[  R(\bm{s},\bm{a}) - \frac{1}{\beta}  \log \frac{\pi(\bm{a}|\bm{s})}{\pi_{\text{marginal}}(\bm{a})}\right] ,
\end{equation}
where $C>0$ is an upper bound on the mutual information between states and actions $I(\bm{S},\bm{A})$ and $\pi_{\text{marginal}}(\bm{a})=\sum_{\bm{s}}p(\bm{s})\pi(\bm{a}|\bm{s})$ is the marginal action distribution. $\beta$ steers the trade-off between reward maximization and mutual information minimization in the unconstrained problem description.

The intuition behind the above objective is to interpret an agent as an information-theoretic channel $\pi(\bm{a}|\bm{s})$ with input $\bm{s}$ and output $\bm{a}$. The agent aims at reconstructing the input at the output as described by the function $R(\bm{s},\bm{a})$. Perfect reconstruction would generally incur high mutual information between input and output but the channel capacity $I(\bm{S},\bm{A})$ is upper-bounded via $C$. This requires therefore the agent to discard such information in $\bm{s}$ that has little impact on $R$ in order to not violate the information constraint. The framework is equivalent to rate distortion theory from information theory~\cite{Cover2006,Genewein2015} (a special case of the information bottleneck~\cite{Tishby1999}) and is versatile in the sense that it applies to a wide range of decision-making systems. In the past, it has been applied to different scientific fields such as: A) economics---where it is referred to as \emph{rational inattention}~\cite{Sims2003}---describing humans as bounded-rational decision-makers; B) information-theoretic decision making to explain the emergence of abstract representations as a consequence of limited information-processing capabilities~\cite{Genewein2015}; C) theoretical neuroscience to develop biologically plausible weight update rules for spiking neurons that prevent synaptic growth without bounds~\cite{Leibfried2015}; and D) machine learning where it translates into a regularizer improving generalization in deep neural networks in classification tasks~\cite{Leibfried2016}.


\section{Mutual-Information Regularization in MDPs}
\label{sec:mdp}

In the next two sections, we first formulate the problem of mutual-information regularization in MDPs from an inference perspective following~\cite{Levine2018,Grau-Moya2019}---see Section~\ref{sec:problem_formulation}. We then proceed by providing a theoretical analysis of the corresponding Bellman operator in Section~\ref{sec:theorie}.

\subsection{Problem Formulation from an Inference Perspective}
\label{sec:problem_formulation}

In statistical inference, the goal is to infer the distribution of some latent variables given observations. The connection between latent and observable variables is determined through a generative model that specifies a prior distribution over latent variables and a likelihood of observable variables conditioned on latent variables. Interestingly, the RL problem of maximizing expected cumulative reward can be phrased as an inference problem by specifying latents and observables in a certain way~\cite{Levine2018} as outlined in the following. Assuming an undiscounted finite horizon problem with horizon $T$, latents are specified as the state-action trajectory $\tau = \bm{s}_0, \bm{a}_0, ..., \bm{s}_T, \bm{a}_T$ distributed according to the prior distribution $p(\tau) = p_{\text{ini}}(\bm{s}_0) \pi_{\text{prior}}(\bm{a}_0) \prod_{t=1}^T P(\bm{s}_t|\bm{s}_{t-1},\bm{a}_{t-1}) \pi_{\text{prior}}(\bm{a}_t)$ where $p_{\text{ini}}: \mathcal{S} \rightarrow [0;1]$ is an initial state distribution and $\pi_{\text{prior}}: \mathcal{A} \rightarrow [0;1]$ is a prior action distribution that is state-unconditioned. A single observable is specified artificially as a binary variable $r$ distributed according to the likelihood $p(r=1|\tau) = \frac{1}{Z} \exp \left( \beta \sum_{t=0}^T R(\bm{s}_t,\bm{a}_t) \right)$ where $Z$ is a normalization constant and $\beta > 0$ a scaling factor. The inference goal is to identify the posterior distribution $p(\tau|r=1) = \frac{p(r=1|\tau) p(\tau)}{\sum_{\tau}p(r=1|\tau) p(\tau)}$ over trajectories given the observation $r=1$. Since computing this posterior is in general intractable, common practice is to resort to variational inference and approximate it via a variational distribution $q(\tau)= p_{\text{ini}}(\bm{s}_0) \pi(\bm{a}_0|\bm{s}_0) \prod_{t=1}^T P(\bm{s}_t|\bm{s}_{t-1},\bm{a}_{t-1}) \pi(\bm{a}_t|\bm{s}_t)$ that is parameterized via the policy $\pi: \mathcal{S} \times \mathcal{A} \rightarrow [0;1]$, a state-conditioned probability distribution. An optimal $q$ implied by an optimal policy $\pi$ is then identified through maximizing a lower bound of the log marginal likelihood $\log \sum_{\tau} p(r=1|\tau) p(\tau)$ referred to as the evidence lower bound (ELBO):
\begin{equation}
\label{eq:elbo}
\begin{split}
\log \sum_{\tau} p(r=1|\tau) p(\tau) & \geq \max_{q} \sum_{\tau} q(\tau) \left( \log p(r=1|\tau) - \log \frac{q(\tau)}{p(\tau)} \right) \\
& \equiv \max_{\pi} \mathbb{E}_{p_{\text{ini}},\pi,P} \left[ \sum_{t=0}^T R(\bm{s}_t,\bm{a}_t) - \frac{1}{\beta} \log \frac{\pi(\bm{a}_t|\bm{s}_t)}{\pi_{\text{prior}}(\bm{a}_t)} \right] ,
\end{split}
\end{equation}
where the `$\equiv$'-sign is because of re-scaling by $\frac{1}{\beta}$ and ignoring the normalization constant $Z$ in $p(r=1|\tau)$ since it only adds a constant offset to the objective. Equation~\eqref{eq:elbo} demonstrates how to recover the RL problem under a soft policy constraint~\cite{Ziebart2010,Azar2011,Fox2016,Haarnoja2017,Nachum2017,Neu2017,Schulman2017,Haarnoja2018,Leibfried2018,Haarnoja2019} from an inference perspective. More details on the subject can be found for example in~\cite{Levine2018}.

Common practice in contemporary variational inference methods is to optimize the ELBO not only w.r.t. to the variational distribution $q$ but also w.r.t. aspects of the generative model itself~\cite{Hoffman2013}---for example the prior---in order to obtain a better log marginal likelihood. Following this reasoning suggests to \emph{optimize} over the prior action distribution $\pi_{\text{prior}}$ rather than keeping it fixed, leading to
\begin{equation}
\label{eq:optimal_elbo}
\max_{\pi,\pi_{\text{prior}}} \mathbb{E}_{p_{\text{ini}},\pi,P} \left[ \sum_{t=0}^T R(\bm{s}_t,\bm{a}_t) - \frac{1}{\beta} \log \frac{\pi(\bm{a}_t|\bm{s}_t)}{\pi_{\text{prior}}(\bm{a}_t)} \right] ,
\end{equation}
which is similar to Equation~\eqref{eq:elbo} except for optimizing over the state-unconditioned prior as well.

Notice how Equation~\eqref{eq:optimal_elbo} resembles the non-sequential decision-making formulation with mutual-information regularization from Equation~\eqref{eq:rationa_inattention} in Section~\ref{sec:rational_inattention}. What remains to be understood is how the incurred logarithmic penalty signal relates to a mutual information constraint. We proceed in line with~\cite{Grau-Moya2019} by expressing the unnormalized marginal state distribution $p^{\pi}(\bm{s})$ of the policy $\pi$ as
\begin{equation}
\label{eq:marginal_state_dis}
p^{\pi}(\bm{s}) = p_{\text{ini}}(\bm{s}) +\sum_{t=1}^T p^{\pi}_t(\bm{s}) \; \text{with} \; p^{\pi}_t(\bm{s}) = \sum_{\bm{s}_0, ..., \bm{a}_{t-1}} p_{\text{ini}}(\bm{s}_0) \prod_{t^{\prime}=1}^t \pi(\bm{a}_{t^{\prime}-1}|\bm{s}_{t^{\prime}-1}) P(\bm{s}_{t^{\prime}}|\bm{s}_{t^{\prime}-1},\bm{a}_{t^{\prime}-1}) ,
\end{equation}
where $\bm{s} \widehat{=} \bm{s}_t$ for $p^{\pi}_t(\bm{s})$. Optimizing Equation~\eqref{eq:optimal_elbo} w.r.t. $\pi_{\text{prior}}$ for a fixed $\pi$ can then be formulated as
\begin{equation}
\label{eq:sequential_mi}
\max_{\pi_{\text{prior}}} \mathbb{E}_{p^{\pi},\pi} \left[ R(\bm{s},\bm{a}) - \frac{1}{\beta} \log \frac{\pi(\bm{a}|\bm{s})}{\pi_{\text{prior}}(\bm{a})} \right] \; \; \Longleftrightarrow \; \; \min_{\pi_{\text{prior}}} \mathbb{E}_{p^{\pi}} \left[ D_{KL} \left( \pi(\cdot|\bm{s}) \middle\Vert \pi_{\text{prior}}(\cdot) \right) \right],
\end{equation}
which corresponds to minimizing the expected Kullback-Leibler divergence between the policy $\pi$ and the prior $\pi_{\text{prior}}$ averaged over the unnormalized marginal state distribution $p^{\pi}$. It is known that the optimal solution to this problem is just the marginal policy $\pi^{\star}_{\text{prior}} (\bm{a}) = \sum_{\bm{s}} \pi(\bm{a}|\bm{s}) p^{\pi}_{\text{norm}}(\bm{s}) = \pi_{\text{marginal}}(\bm{a})$ where $p^{\pi}_{\text{norm}}$  refers to the normalized marginal state distribution~\cite{Cover2006, Grau-Moya2019}. Replacing $\pi_{\text{prior}}$ in the expected $D_{KL}$-term in Equation~\eqref{eq:sequential_mi} with $\pi_{\text{marginal}}$ yields the unnormalized mutual information $I(\bm{S},\bm{A})$ scaled by a factor as a consequence of the unnormalized marginal state distribution. Utilizing the marginal policy $\pi_{\text{marginal}}$, we can express Equation~\eqref{eq:optimal_elbo} finally as follows
\begin{equation}
\label{eq:optimal_elbo_ref}
\max_{\pi} \mathbb{E}_{p^{\pi},\pi} \left[ R(\bm{s},\bm{a}) - \frac{1}{\beta} \log \frac{\pi(\bm{a}|\bm{s})}{\pi_{\text{marginal}}(\bm{a})} \right] \; \; \Longleftrightarrow \; \; \max_{\pi} \mathbb{E}_{p^{\pi},\pi} \left[ R(\bm{s},\bm{a}) \right] \; \text{s.t.} \; I(\bm{S},\bm{A}) \leq C,
\end{equation}
in its unconstrained and constrained form respectively bridging the gap to the non-sequential formulation in Equation~\eqref{eq:rationa_inattention} from Section~\ref{sec:rational_inattention}. Importantly, Equation~\eqref{eq:optimal_elbo_ref} is also valid in a discounted infinite horizon setting where $p^{\pi}(\bm{s}) = p_{\text{ini}}(\bm{s}) + \sum_{t=1}^{\infty} \gamma^t p^{\pi}_t(\bm{s})$ as explained for example in~\cite{Grau-Moya2019}.

A notable difference compared to the non-sequential setting is that the unnormalized marginal state distribution $p^{\pi}$ depends on the optimization argument $\pi$, which renders the optimization problem non-trivial. It has therefore been suggested to introduce a policy-independent state distribution $p: \mathcal{S} \rightarrow [0;1]$~\cite{Tishby2011} for computing the marginal policy $\pi_{\text{marginal}}$---stronger consistency assumptions on $p$ that adhere to state transitions have not been investigated so far and are left for future work. For infinite horizon scenarios, we then obtain the following Bellman operator:
\begin{equation}
\label{eq:value_iteration}
B_{\star}V(\bm{s}) := \max_\pi \mathbb{E}_{\pi} \left[ R(\bm{s},\bm{a}) - \frac{1}{\beta} \log \frac{\pi(\bm{a}|\bm{s})}{\pi_{\text{prior}}(\bm{a})} + \gamma \mathbb{E}_{P}\left[ V(\bm{s}^{\prime}) \right] \right] \; \text{s.t.} \; \pi_{\text{prior}}(\bm{a}) = \sum_{\bm{s}} \pi(\bm{a}|\bm{s}) p(\bm{s}) ,
\end{equation}
where the constrained problem formulation highlights that the optimal prior $\pi^{\star}_{\text{prior}}$ is the true marginal action distribution $\pi_{\text{marginal}}(\bm{a})=\sum_{\bm{s}} \pi(\bm{a}|\bm{s}) p(\bm{s})$ as outlined earlier. The marginal action distribution however depends on the behavioral policy in all states, not just the one that occurs above on the l.h.s. of the equation in $B_{\star}V(\bm{s})$. This leads to the problem that the Bellman operator $B_{\star}: \mathbb{R}^{\mathcal{S}} \rightarrow \mathbb{R}^{\mathcal{S}}$ cannot be applied independently to all states $\bm{s} \in \mathcal{S}$ as in ordinary value iteration schemes. The consequence of the latter is that standard theoretical tools for analyzing unique optimal values and convergence cannot be applied straightforwardly~\cite{Bertsekas1996,Rubin2012,Grau-Moya2016}, although there is empirical evidence that mutual-information regularized value iteration can convergence in grid world examples~\cite{Tishby2011}.

Apart from the difficulty of analyzing the overall convergence behaviour of mutual-information regularized value iteration, applying the actual Bellman operator $B_{\star}$ from Equation~\eqref{eq:value_iteration} is also not trivial since there is no closed-form solution for an optimal policy-prior pair. In the next section (Section~\ref{sec:theorie}), we therefore provide a theoretical analysis on how to \emph{apply} the operator $B_{\star}$ to a given value function $V: \mathcal{S} \rightarrow \mathbb{R}$. This results in a practical algorithm that iteratively re-computes the optimal `one-step' policy and the optimal `one-step' prior w.r.t. the current value estimates in an alternate fashion until convergence. In Section~\ref{sec:rl}, we devise a novel actor-critic algorithm inspired by Equation~\eqref{eq:value_iteration} for continuous action spaces, and demonstrate that it can attain competitive performance with state-of-the-art algorithms in the robotics simulation domain of Mujoco.

\textbf{Remark 1} \textit{In a broader sense, the formulation above relates to other sequential decision-making formulations based on the information bottleneck~\cite{Goyal2019} and information-processing hierarchies~\cite{Hihn2019}.}

\subsection{Theoretical Analysis of the Bellman Operator}
\label{sec:theorie}

Our theoretical contribution (Theorem~\ref{theo:blahut-arimoto}) is an iterative algorithm to apply the Bellman operator $B_{\star}$.

\begin{theorem}{On the Application of the Bellman Operator $B_{\star}$.}
\label{theo:blahut-arimoto}
Under the assumption that the reward function $R$ is bounded, the optimization problem in Equation~(\ref{eq:value_iteration}) imposed by the Bellman operator $B_{\star}$ can be solved by iterating in an alternate fashion through the following two equations:
\begin{equation}
\label{eq:optimal_prior}
\pi_{\text{\emph{prior}}}^{(m)}(\bm{a}) = \sum_{\bm{s}} \pi^{(m)}(\bm{a}|\bm{s}) p(\bm{s}),
\end{equation}
\begin{equation}
\label{eq:optimal_policy}
\pi^{(m+1)}(\bm{a}|\bm{s}) = \frac{\pi_{\text{\emph{prior}}}^{(m)}(\bm{a}) e^{ \beta \left( R(\bm{s},\bm{a}) + \gamma \sum_{\bm{s}^{\prime}} P(\bm{s}^{\prime}|\bm{s},\bm{a}) V(\bm{s}^{\prime}) \right)}}{\sum_{\bm{a}} \pi_{\text{\emph{prior}}}^{(m)}(\bm{a}) e^{ \beta \left( R(\bm{s},\bm{a}) + \gamma \sum_{\bm{s}^{\prime}} P(\bm{s}^{\prime}|\bm{s},\bm{a}) V(\bm{s}^{\prime}) \right)}},
\end{equation}
where $m$ refers to the iteration index. Denoting the total number of iterations as $M$, the presented scheme converges at a rate of $O(1/M)$ to an optimal policy $\pi^{\star}$ for any given bounded value function $V$ and any initial policy $\pi^{(0)}:\mathcal{S} \times  \mathcal{A} \rightarrow [0;1]$ that has support in $\mathcal{A} \; \forall \; \bm{s} \in \mathcal{S}$. Such a scheme is commonly referred to as Blahut-Arimoto-type algorithm~\cite{Cover2006}.
\end{theorem}

\textbf{Proof.} The problem is similar to rate distortion theory~\cite{Cover2006} from information theory and so is identifying an optimal solution, accomplished via Lemmas~\ref{lemma:optimal_prior} and~\ref{lemma:optimal_policy} and Proposition~\ref{prop:convergence} next. Lemma~\ref{lemma:optimal_prior} deals with how to compute an optimal prior given a fixed policy, while Lemma~\ref{lemma:optimal_policy} deals with how to compute an optimal policy given a fixed prior. This leads to a set of self-consistent equations whose alternate application converges to an optimal solution as proven by Proposition~\ref{prop:convergence}. \hfill $\square$

\begin{definition}{Evaluation Operator.}
A specific prior-policy-pair $\left(\pi_{\text{\emph{prior}}}, \pi \right)$ is evaluated as
\label{def:eval_op}
\begin{equation}
\label{eq:evaluation_operator}
B_{\pi_{\text{\emph{prior}}},\pi} V(\bm{s}) := \mathbb{E}_{\pi} \left[ R(\bm{s},\bm{a}) - \frac{1}{\beta} \log \frac{\pi(\bm{a}|\bm{s})}{\pi_{\text{\emph{prior}}}(\bm{a})} + \gamma \mathbb{E}_{P}\left[ V(\bm{s}^{\prime}) \right] \right] .
\end{equation}
\end{definition}

\begin{lemma}{Optimal Prior for a Given Policy.}
\label{lemma:optimal_prior}
The optimal prior for Equation~(\ref{eq:value_iteration}) given a bounded reward function $R$, a bounded value function $V$ and a policy $\pi$ is:
\begin{equation}
\label{eq:optimal_prior_2}
\argmax_{\pi_{\text{\emph{prior}}}} \mathbb{E}_{p} \left[ B_{\pi_{\text{\emph{prior}}},\pi} V(\bm{s}) \right] = \sum_{\bm{s}} \pi(\bm{a}|\bm{s}) p(\bm{s}) .
\end{equation}
\end{lemma}

\textbf{Proof.} Maximizing the expected evaluation operator $\mathbb{E}_{p} \left[ B_{\pi_{\text{prior}},\pi} V(\bm{s}) \right]$ w.r.t. $\pi_{\text{prior}}$ is equivalent to minimizing the expected Kullback-Leibler divergence $\mathbb{E}_{p} \left[ D_{KL} \left( \pi(\cdot|\bm{s}) \Vert \pi_{\text{prior}}(\cdot) \right) \right]$, because the reward and the value function do not depend on the prior. It holds that $I(\bm{S},\bm{A}) \leq \mathbb{E}_{p} \left[ D_{KL} \left( \pi(\cdot|\bm{s}) \Vert \pi_{\text{prior}}(\cdot) \right) \right]$ for all $\pi_{\text{prior}}$~\cite{Cover2006}, implying that the optimal prior is the true marginal action distribution averaged over all states. \hfill $\square$

\begin{lemma}{Optimal Policy for a Given Prior.}
\label{lemma:optimal_policy}
The optimal policy for Equation~(\ref{eq:value_iteration}) given a bounded reward function $R$, a bounded value function $V$ and a prior policy $\pi_{\text{\emph{prior}}}$ is:
\begin{equation}
\label{eq:optimal_policy_2}
\argmax_{\pi}  B_{\pi_{\text{\emph{prior}}},\pi} V(\bm{s}) = \frac{\pi_{\text{\emph{prior}}}(\bm{a}) e^{\beta \left( R(\bm{s},\bm{a}) + \gamma \sum_{\bm{s}^{\prime}} P(\bm{s}^{\prime}|\bm{s},\bm{a}) V(\bm{s}^{\prime}) \right)}}{\sum_{\bm{a}} \pi_{\text{\emph{prior}}}(\bm{a}) e^{ \beta \left( R(\bm{s},\bm{a}) + \gamma \sum_{\bm{s}^{\prime}} P(\bm{s}^{\prime}|\bm{s},\bm{a}) V(\bm{s}^{\prime}) \right)}} .
\end{equation}
\end{lemma}

\textbf{Proof.} First note that for a given prior $\pi_{\text{prior}}$, Equation~\eqref{eq:value_iteration} can be solved independently for each state $\bm{s} \in \mathcal{S}$. Identifying an optimal policy then becomes an optimization problem subject to the constraint $\sum_{\bm{a}}\pi(\bm{a}|\bm{s})=1$, which can be straightforwardly solved with the method of Lagrange multipliers and standard variational calculus yielding Equation~\eqref{eq:optimal_policy_2}---see Appendix~\ref{sec:details_policy_derivation} in line with~\cite{Genewein2015}. \hfill $\square$

\begin{corollary}{Concise Bellman Operator.}
\label{cor:certainty_equivalent}
For a specific prior-policy-pair $\left(\pi_{\text{\emph{prior}}}^{(m)}, \pi^{(m+1)}\right)$ obtained by running the Blahut-Arimoto scheme for $m$ rounds according to Equations~(\ref{eq:optimal_prior}) and~(\ref{eq:optimal_policy}), it holds:
\begin{equation}
\label{eq:certainty_equivalent}
B_{\pi^{(m)}_{\text{\emph{prior}}},\pi^{(m+1)}} V(\bm{s}) = \frac{1}{\beta} \log \mathbb{E}_{\pi_{\text{\emph{prior}}}^{(m)}} \left[ e^{ \beta \left( R(\bm{s},\bm{a}) + \gamma \mathbb{E}_P \left[ V(\bm{s}^{\prime}) \right]\right) } \right],
\end{equation}
which is obtained by plugging Equation~(\ref{eq:optimal_policy}) into $B_{\pi_{\text{\emph{prior}}},\pi} V(\bm{s})$ from Equation~(\ref{eq:evaluation_operator}) under the assumption of having the `compatible' prior $\pi_{\text{\emph{prior}}}=\pi^{(m)}_{\text{\emph{prior}}}$.
\end{corollary}

\begin{proposition}{Convergence.}
\label{prop:convergence}
Given a bounded reward function $R$, a bounded value function $V$ and an initial policy $\pi^{(0)}$ with support in $\mathcal{A} \; \forall \; \bm{s} \in \mathcal{S}$, iterating through Equations~(\ref{eq:optimal_prior}) and~(\ref{eq:optimal_policy}) converges to an optimal policy $\pi^{\star}$ at a rate of $O(1/M)$ where $M$ is the total number of iterations.
\end{proposition}

\textbf{Proof.} The proof is accomplished via Lemmas~\ref{lemma:upper_value_bound} and~\ref{lemma:completing_convergence} in line with~\cite{Gallager1994}. We first specify an upper bound (Lemma~\ref{lemma:upper_value_bound}) for the solution to the optimization problem in Equation~\eqref{eq:value_iteration}, i.e. the expected Bellman operator averaged over states, which we use to complete the proof in Lemma~\ref{lemma:completing_convergence}. \hfill $\square$

\begin{lemma}{Upper Expected Optimal Value Bound.}
\label{lemma:upper_value_bound}
The solution to Equation~(\ref{eq:value_iteration}) is upper-bounded:
\begin{equation}
\label{eq:upper_value_bound}
\mathbb{E}_{p} \left[ B_{\star} V(\bm{s}) \right] \leq \frac{1}{\beta} \mathbb{E}_{p,\pi^{\star}} \left[ \log \frac{\pi^{(m+1)}(\bm{a}|\bm{s})}{\pi^{(m)}(\bm{a}|\bm{s})} \right] + \mathbb{E}_{p} \left[ B_{\pi^{(m)}_{\text{\emph{prior}}},\pi^{(m+1)}} V(\bm{s}) \right] ,
\end{equation}
where the superscript $^{\star}$ refers to an optimal solution, and the superscripts $^{(m)}$ and $^{(m+1)}$ to preliminary solutions after $m$ and $m+1$ iterations of the Blahut-Arimoto scheme respectively.
\end{lemma}

\textbf{Proof.} The key ingredient for deriving the upper bound in Equation~\eqref{eq:upper_value_bound} is to show that for all $m$:
\begin{equation}
\label{eq:upper_value_bound_2}
\mathbb{E}_p \left[ B_{\star}V(\bm{s}) \right] \leq \mathbb{E}_{p,\pi^{\star}} \left[ R(\bm{s},\bm{a}) - \frac{1}{\beta} \log \frac{\pi^{(m)}(\bm{a}|\bm{s})}{\pi^{(m)}_{\text{prior}}(\bm{a})} + \gamma \mathbb{E}_{P}\left[ V(\bm{s}^{\prime}) \right] \right] ,
\end{equation}
which we detail in the first part of Appendix~\ref{sec:details_upper_value_bound} with help of~\cite{Polyanskiy2016}. Using Equation~\eqref{eq:optimal_policy} for $\pi^{(m+1)}$ and Corollary~\ref{cor:certainty_equivalent}, we then obtain Equation~\eqref{eq:upper_value_bound} as detailed in the second part of Appendix~\ref{sec:details_upper_value_bound}. \hfill $\square$

\begin{lemma}{Completing Convergence.}
\label{lemma:completing_convergence}
The convergence proof is completed by showing that
\begin{equation}
\label{eq:completing_convergence}
\frac{1}{M} \sum_{m=0}^{M-1} \left( \mathbb{E}_{p} \left[ B_{\star} V(\bm{s}) \right] - \mathbb{E}_{p} \left[ B_{\pi^{(m)}_{\text{\emph{prior}}},\pi^{(m+1)}} V(\bm{s}) \right]  \right) \leq \frac{1}{M \beta} \mathbb{E}_{p} \left[ \max_{\bm{a}} \log \left( \pi^{(0)} (\bm{a}|\bm{s}) \right)^{-1} \right] ,
\end{equation}
where $\pi^{(0)}$ is the initial policy at the start of the Blahut-Arimoto algorithm, implying a rate $O(1/M)$.
\end{lemma}

\textbf{Proof.} This is obtained by rearranging Equation~\eqref{eq:upper_value_bound} such that only the logarithmic policy term remains on the right-hand side. Taking the average over $M$ iterations and swapping the expectation over $p$ and $\pi^{\star}$ with the sum on the right-hand side, all the logarithmic policy terms except for $M$ and $0$ cancel. Under worst case assumptions, one arrives at Equation~\eqref{eq:completing_convergence}---details in Appendix~\ref{sec:details_convergence_proof}. \hfill $\square$

\begin{figure}[ht]
\begin{center}
\centerline{\includegraphics[width=0.7\columnwidth]{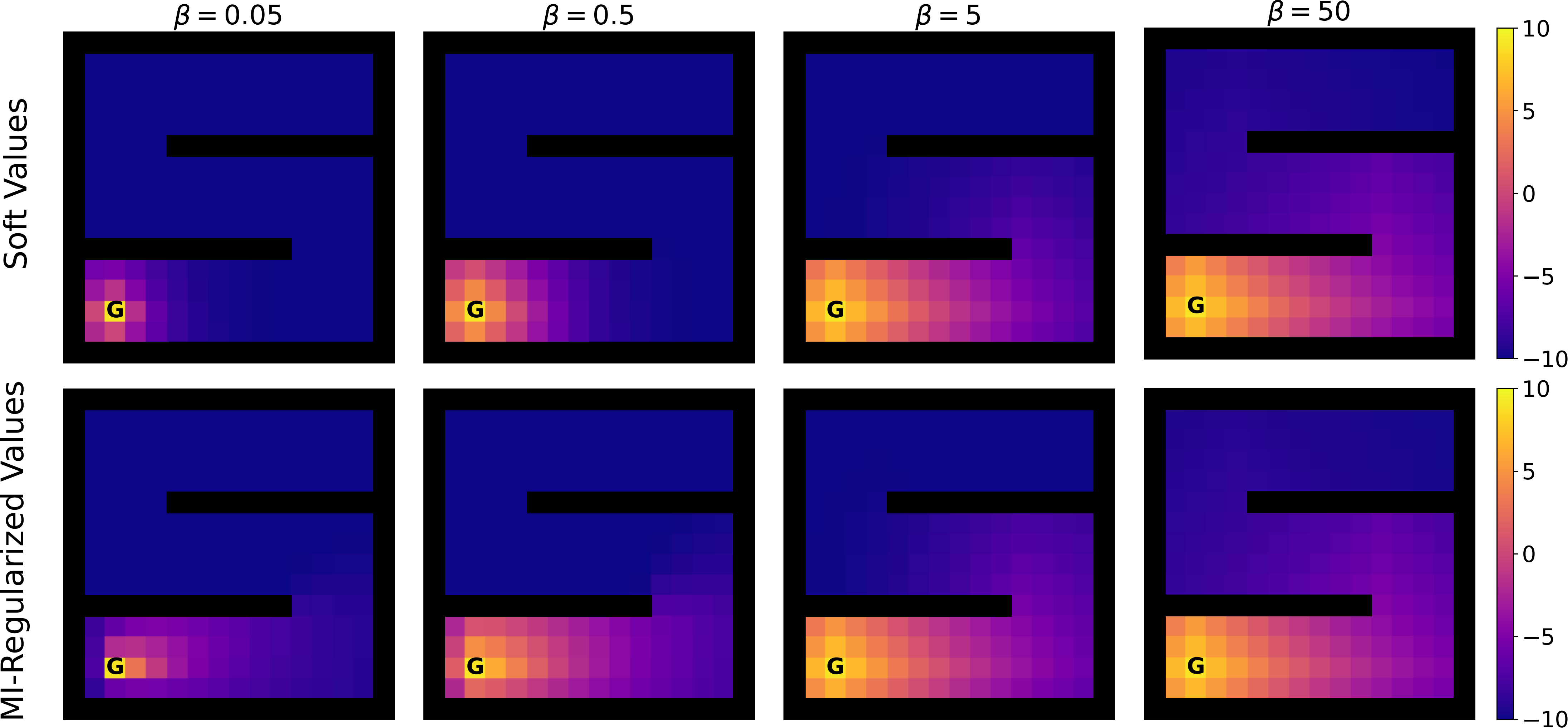}}
\caption{Effect of $\beta$ on State Values. Soft values (uniform prior) and mutual-information-regularized values are compared. For small $\beta$, soft values evaluate the prior while mutual-information-regularized values are optimal for a state-independent policy---details in Appendix~\ref{sec:grid_world}.}
\label{fig:grid}
\end{center}
\end{figure}

In Figure~\ref{fig:grid}, we validate $B_\star$ in a grid world confirming that mutual-information regularized value iteration can converge in line with~\cite{Tishby2011}, and compare against soft values where the prior is fixed.

\textbf{Remark 2} \textit{Corollary~\ref{cor:certainty_equivalent} shows that the mutual-information regularized Bellman operator generalizes the soft Bellman operator~\cite{Azar2011,Rubin2012,Fox2016,Neu2017,Schulman2017,Leibfried2018}, and cumulative entropy regularization as a further special case of the latter~\cite{Ziebart2010,Haarnoja2017,Nachum2017,Haarnoja2018,Levine2018,Haarnoja2019}, when fixing the prior and not optimizing it.}

\section{Mutual-Information Regularization with Continuous Actions}
\label{sec:rl}

While the previous section (Section~\ref{sec:theorie}) deals with a theoretical analysis of the Bellman operator $B_{\star}$ resulting from mutual-information regularization, we devise here a mutual-information regularized actor-critic learning (MIRACLE) algorithm inspired by $B_{\star}$ that scales to high dimensions and can handle continuous action spaces. Our hypothesis is that optimizing the ELBO from Equation~\eqref{eq:elbo} w.r.t. the prior $\pi_{\text{prior}}$ should lead to a better log marginal likelihood~\cite{Hoffman2013} and hence to better performance, as empirically verified with deep parametric function approximators in domains with discrete actions such as Atari~\cite{Grau-Moya2019}. In the next section (Section~\ref{sec:result}), we empirically validate our method in the robotics simulation domain Mujoco and demonstrate competitive performance with contemporary methods. Specifically, we show that optimizing over the prior can improve over the state-of-the-art soft actor-critic (SAC) algorithm~\cite{Haarnoja2018,Haarnoja2019} that implicitly assumes a fixed uniform prior.

Following the policy gradient theorem~\cite{Degris2012} and in line with Equation~\eqref{eq:actor_critic}, MIRACLE optimizes the parameters $\phi$ of a parametric policy $\pi_{\phi}: \mathcal{S} \times \mathcal{A} \rightarrow [0;1]$ by gradient ascent on the objective:
\begin{equation}
\label{eq:parametric_policy}
J_{\phi} = \mathbb{E}_{\mathcal{D} ,\pi_\phi} \left[ Q_{\theta}(\bm{s},\bm{a}) - \frac{1}{\beta} \log \frac{\pi_{\phi}(\bm{a}|\bm{s})}{\pi_{\chi}(\bm{a})} \right] ,
\end{equation}
where states are sampled from an empirical non-stationary distribution, i.e. a replay buffer $\mathcal{D}$ that collects environment interactions of the agent~\cite{Lillicrap2016,Abdolmaleki2018,Fujimoto2018,Haarnoja2018}, and actions from the policy $\pi_{\phi}$. $Q_{\theta}: \mathcal{S} \times \mathcal{A} \rightarrow \mathbb{R}$ and $\pi_{\chi}: \mathcal{A} \rightarrow [0;1]$ are additional function approximators with parameters $\theta$ and $\chi$ that approximate the Q-values of the policy $\pi_{\phi}$ and the marginal policy of $\pi_{\phi}$ averaged over states respectively. The logarithmic term that penalizes deviations of the policy $\pi_{\phi}$ from its marginal $\pi_{\chi}$ is a direct consequence of mutual-information regularization, where we leverage on the fact that the optimal prior state-unconditioned policy is the true marginal action distribution---see Section~\ref{sec:problem_formulation}.

To learn Q-values efficiently, we follow~\cite{Haarnoja2018} and introduce an additional V-critic $V_{\psi}: \mathcal{S} \rightarrow \mathbb{R}$ parameterized by $\psi$ to approximate state values of the policy $\pi_{\phi}$. Optimal parameters $\theta$ of the Q-critic are then learned via gradient descent on the following average squared loss:
\begin{equation}
\label{eq:parametric_q_critic}
J_{\theta} = \mathbb{E}_\mathcal{D} \left[ \left( Q_{\theta}(\bm{s},\bm{a}) - R(\bm{s},\bm{a}) - \gamma V_{\psi}(\bm{s}^{\prime}) \right)^2 \right] ,
\end{equation}
where $\mathcal{D}$ refers to a distribution that samples state-action-next-state tuples from the replay buffer.

Optimal parameters of the V-critic are trained via gradient descent on the following loss:
\begin{equation}
\label{eq:parametric_v_critic}
J_{\psi} = \mathbb{E}_\mathcal{D} \left[ \left( V_{\psi}(\bm{s}) - \mathbb{E}_{\pi_{\phi}} \left[ Q_{\theta}(\bm{s},\bm{a}) - \frac{1}{\beta} \log \frac{\pi_{\phi}(\bm{a}|\bm{s})}{\pi_{\chi}(\bm{a})} \right] \right)^2 \right] ,
\end{equation}
where the replay buffer $\mathcal{D}$ is used to sample states and the policy $\pi_{\phi}$ to sample actions.

Finally, one needs to approximate the (state-unconditioned) marginal action distribution $\pi_{\chi}$ of the policy $\pi_{\phi}$. Abusing notation, we learn a probabilistic parametric map $\pi_{\chi}: \mathbb{R}^{\dim(\mathcal{A})} \times \mathcal{A} \rightarrow [0;1]$, i.e. a conditional probability distribution $\pi_{\chi}(\bm{a}|\bm{u})$, that maps samples $\bm{u} \sim \mathcal{N} \left( \bm{u} \middle| \bm{0}_{\dim(\mathcal{A})}, \bm{I}_{\dim(\mathcal{A})} \right)$ from the standard Gaussian to actions $\bm{a} \in \mathcal{A}$ from the replay buffer $\mathcal{D}$. This can be phrased as a supervised problem and achieved with max log likelihood.
The parametric marginal action distribution is then: $\pi_{\chi}(\bm{a})=\int_{\bm{u}}\pi_{\chi}(\bm{a}|\bm{u})\mathcal{N} \left( \bm{u} \middle| \bm{0}_{\dim(\mathcal{A})}, \bm{I}_{\dim(\mathcal{A})} \right)d \bm{u}$ and can be approximated  as $\pi_{\chi}(\bm{a}) \approx \frac{1}{N} \sum_{n=1}^N \pi_{\chi}(\bm{a}|\bm{u}_n)$ to estimate probability values in Equations~\eqref{eq:parametric_policy} and~\eqref{eq:parametric_v_critic}, where $\bm{u}_n$ are i.i.d. Gaussian samples. A fixed state distribution, as in our theoretical analysis in Section~\ref{sec:theorie} and in~\cite{Tishby2011} for tabular environments, would be non-sensible in Mujoco because of the problem of admissible states. The marginal action distribution marginalizes therefore over states from the replay buffer instead to approximate the policy-induced marginal action distribution as outlined in Section~\ref{sec:problem_formulation} and in line with~\cite{Grau-Moya2019}. Experiments using a variational autoencoder with a recognition model~\cite{Kingma2014} to approximate the marginal policy did not yield better results---see Appendix~\ref{sec:implementation_details} and~\ref{sec:raw_results}.

The training procedure for MIRACLE is off-policy and similar to~\cite{Haarnoja2018} where the agent interacts with the environment iteratively and stores state-action-reward-next-state tuples in a replay buffer. After every step, a minibatch of transition tuples is sampled from the buffer to perform a single gradient step on the objectives for the different learning components of the algorithm. We also practically operate with a reward scale, i.e. we scale the reward function by $\beta$ rather than the logarithmic penalty term by its inverse $\beta^{-1}$~\cite{Haarnoja2018}. During training, we apply the reparameterization trick~\cite{Kingma2014,Rezende2014,Haarnoja2018} for $\pi_{\phi}$ and $\pi_{\chi}$ in their respective objectives---details can be found in Appendix C of~\cite{Haarnoja2018} explaining how to reparameterize when action spaces are bounded (as in Mujoco). 

\section{Experiments in Mujoco}
\label{sec:result}

We empirically validate MIRACLE on the latest v2-environments of Mujoco. Function approximators are parameterized with deep nets trained with Adam~\cite{Kingma2015}. We use a twin Q-critic~\cite{vanHasselt2016,Haarnoja2018} and limit the variance of the probabilistic policy networks $\pi_{\phi}$ and $\pi_{\chi}$ following~\cite{Chua2018}. Further details can be found in Appendix~\ref{sec:implementation_details}. Our baselines are DDPG~\cite{Lillicrap2016} and PPO~\cite{Schulman2017b} from RLlib~\cite{Liang2018} as well as SAC~\cite{Haarnoja2018}. Note that we are mainly concerned comparing to the latter since we introduce a marginal reference policy to be optimized over compared to an implicit fixed uniform marginal policy in SAC---we therefore use an implementation for SAC and MIRACLE that only differs in this aspect and is the same otherwise. Every experiment is conducted with \emph{ten} seeds. 

\begin{figure}[ht]
\begin{center}
\centerline{\includegraphics[width=1.0\columnwidth]{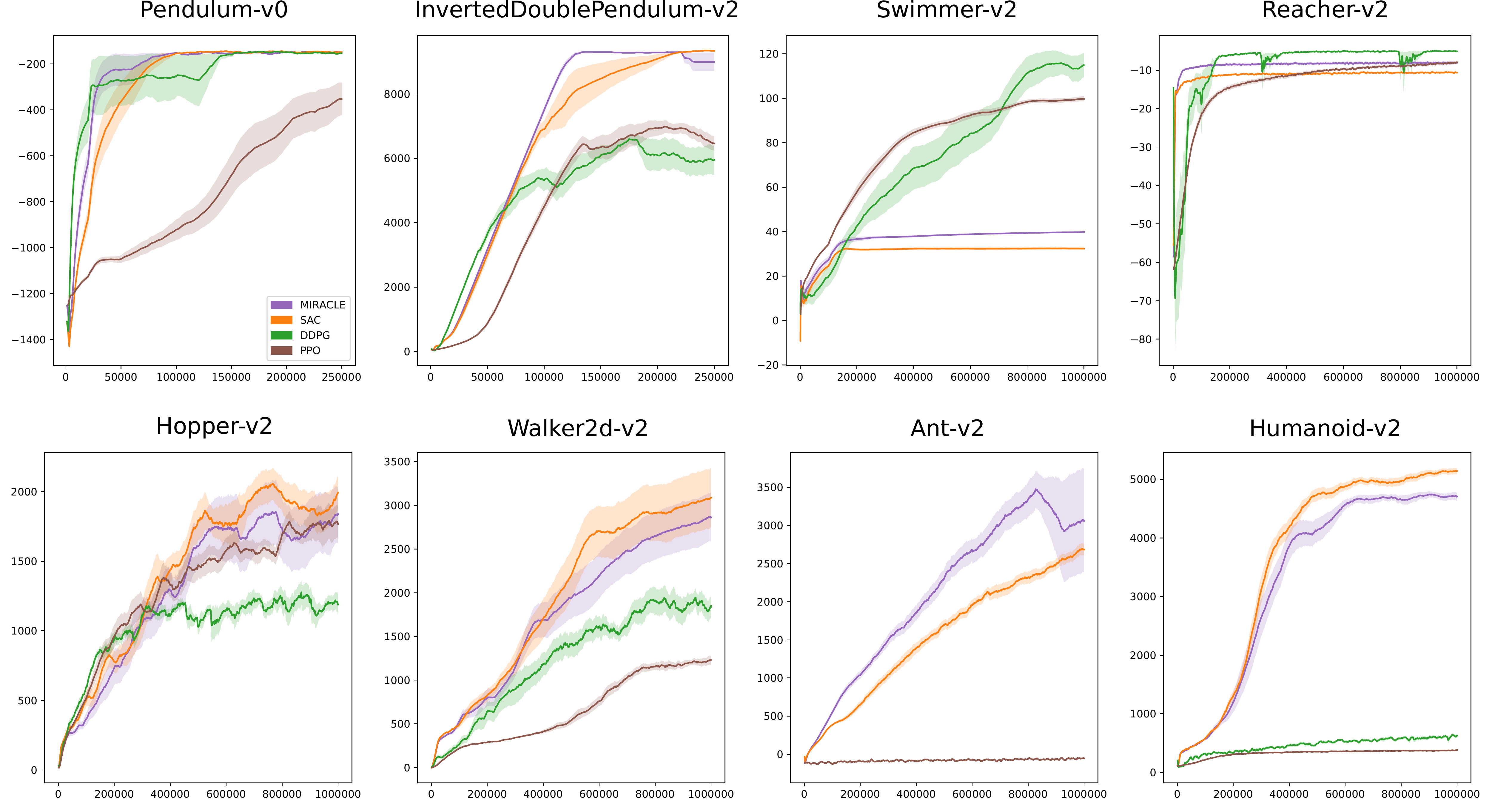}}
\caption{Mujoco Experiments. The figure reports average episodic rewards over the last $100$ episodes and standard errors. MIRACLE consistently outperforms SAC in lower-dimensional environments (top row). In high-dimensional tasks, MIRACLE outperforms DDPG and PPO, and can outperform SAC significantly, see Ant (bottom row). In general, MIRACLE may not always help since it encourages the policy to stay close to past actions, which can be different to actions identified by the optimization procedure in line with experiments in discrete-action environments~\cite{Grau-Moya2019}. In Ant (kink caused by one seed), DDPG from RLlib did not work properly in our setup---see Appendix~\ref{sec:raw_results}.}
\label{fig:best_reward}
\end{center}
\end{figure}

Figure~\ref{fig:best_reward} shows that MIRACLE consistently outperforms SAC on lower-dimensional environments. On high-dimensional environments, MIRACLE is better than DDPG and PPO, and can outperform SAC. However, MIRACLE might not always help since it favours actions that are close to past actions, which may differ from actions that are identified by the optimization procedure~\cite{Grau-Moya2019}.


\section{Conclusion}
\label{sec:conclusion}

Motivating mutual-information-regularization in MDPs from an inference perspective leads to a Bellman operator that generalizes the soft Bellman operator from the literature. We provide a theoretical analysis of this operator resulting in a practically applicable algorithm. Inspired by that, we devise an actor-critic algorithm (with an adaptive marginal prior) for high-dimensional continuous domains and demonstrate competitive results compared to contemporary methods in Mujoco, e.g. improvements in Ant over the state-of-the-art SAC (with an implicit fixed marginal prior).


\acknowledgments{We thank Joshua Aduol for helpful suggestions on the engineering side.}

\bibliography{miraclebib}

\clearpage
\newpage
\appendix

\section{Details Regarding the Proof of Lemma~\ref{lemma:optimal_policy}}
\label{sec:details_policy_derivation}

Formulating the Lagrangian for Lemma~\ref{lemma:optimal_policy} yields:
\begin{equation}
\label{eq:lagrangian}
L(\pi, \lambda) = \mathbb{E}_{\pi} \left[ R(\bm{s},\bm{a}) - \frac{1}{\beta} \log \frac{\pi(\bm{a}|\bm{s})}{\pi_{\text{prior}}(\bm{a})} + \gamma \mathbb{E}_{P}\left[ V(\bm{s}^{\prime}) \right] \right] - \lambda \left( \left( \sum_{\bm{a}} \pi(\bm{a}|\bm{s}) \right) -1 \right) .
\end{equation}
Taking the derivative of $L(\pi, \lambda)$ w.r.t. to $\pi$ for a specific action, equating with zero and resolving for $\pi$ leads to:
\begin{equation}
\label{eq:lagrangian_policy}
\pi_{\text{prior}}(\bm{a}) e^{ \beta \left(R(\bm{s},\bm{a}) + \gamma \mathbb{E}_P \left[ V(\bm{s}^\prime) \right]  \right) -1 -\beta \lambda }.
\end{equation}
Taking the derivative of $L(\pi, \lambda)$ w.r.t. to $\lambda$, plugging in Equation~\eqref{eq:lagrangian_policy}, equating with zero and resolving for $\exp(-1-\beta\lambda)$, one arrives at:
\begin{equation}
\label{eq:lagrangian_normalization}
e^{-1 -\beta \lambda} = \left( \sum_{\bm{a}} \pi_{\text{prior}}(\bm{a}) e^{\beta \left(R(\bm{s},\bm{a}) + \gamma \mathbb{E}_P \left[ V(\bm{s}^\prime) \right]  \right)} \right)^{-1}.
\end{equation}
Plugging Equation~\eqref{eq:lagrangian_normalization} back into Equation~\eqref{eq:lagrangian_policy}, one arrives at Equation~\eqref{eq:optimal_policy_2} from the main paper.

\section{Details Regarding the Proof of Lemma~\ref{lemma:upper_value_bound}}
\label{sec:details_upper_value_bound}

First, we detail how to obtain Equation~\eqref{eq:upper_value_bound_2}. This is accomplished by using the inequality $\mathbb{E}_{p} \left[ D_{KL} \left( \pi^{\star}(\cdot|\bm{s}) \Vert \pi^{(m)}(\cdot|\bm{s}) \right) \right] \geq D_{KL} \left( \pi_{\text{prior}}^{\star}(\cdot) \Vert \pi_{\text{prior}}^{(m)}(\cdot) \right)$---known as `conditioning increases divergence', see e.g.~\cite{Polyanskiy2016}. Rearranging then yields: $\mathbb{E}_{p,\pi^{\star}} \left[ -\log \frac{\pi^{\star}(\bm{a}|\bm{s})}{\pi_{\text{prior}}^{\star}(\bm{a})}\right] \leq \mathbb{E}_{p,\pi^{\star}} \left[ -\log \frac{\pi^{(m)}(\bm{a}|\bm{s})}{\pi_{\text{prior}}^{(m)}(\bm{a})}\right]$, which leads to Equation~\eqref{eq:upper_value_bound_2} because rewards and values are policy-independent. The upper value bound in Equation~\eqref{eq:upper_value_bound} from Lemma~\ref{lemma:upper_value_bound} is then derived starting from Equation~\eqref{eq:upper_value_bound_2} as follows:
\begin{equation}
\label{eq:details_upper_value_bound}
\begin{split}
\mathbb{E}_p \left[ B_{\star}V(\bm{s}) \right] & \leq \mathbb{E}_{p,\pi^{\star}} \left[ R(\bm{s},\bm{a}) - \frac{1}{\beta} \log \frac{\pi^{(m)}(\bm{a}|\bm{s})}{\pi^{(m)}_{\text{prior}}(\bm{a})} + \gamma \mathbb{E}_{P}\left[ V(\bm{s}^{\prime}) \right] \right] \\
& = \mathbb{E}_{p, \pi^{\star}} \left[ \frac{1}{\beta} \log \left( e^{\beta R(\bm{s},\bm{a}) + \log \pi^{(m)}_{\text{prior}} (\bm{a}) + \beta \gamma \mathbb{E}_{P} \left[ V(\bm{s}^{\prime}) \right]} \right) - \frac{1}{\beta} \log \pi^{(m)}(\bm{a}|\bm{s}) \right] \\
& = \mathbb{E}_{p,\pi^{\star}} \left[ \frac{1}{\beta} \log \left( \pi^{(m)}_{\text{prior}} (\bm{a}) e^{\beta \left( R(\bm{s},\bm{a}) + \gamma \mathbb{E}_P \left[ V(\bm{s}^{\prime}) \right] \right)} \right)  - \frac{1}{\beta} \log \pi^{(m)}(\bm{a}|\bm{s}) \right] \\
& = \mathbb{E}_{p,\pi^{\star}} \left[ \frac{1}{\beta} \log \pi^{(m+1)}(\bm{a}|\bm{s}) - \frac{1}{\beta} \log \pi^{(m)}(\bm{a}|\bm{s}) \right] \\
& \; \; \; \; + \mathbb{E}_p \left[\frac{1}{\beta} \log \mathbb{E}_{\pi^{(m)}_{\text{prior}}} \left[e^{\beta \left( R(\bm{s},\bm{a}) + \gamma \mathbb{E}_P \left[ V(\bm{s}^{\prime}) \right] \right)} \right] \right] \\
& = \frac{1}{\beta} \mathbb{E}_{p,\pi^{\star}} \left[ \log \frac{\pi^{(m+1)}(\bm{a}|\bm{s})}{\pi^{(m)}(\bm{a}|\bm{s})} \right] + \mathbb{E}_{p} \left[ B_{\pi^{(m)}_{\text{prior}}, \pi^{(m+1)}} V(\bm{s}) \right] .
\end{split}
\end{equation}
To obtain the second to last line, we use Equation~\eqref{eq:optimal_policy} for $\pi^{(m+1)}$, and to obtain the last line, we use Equation~\eqref{eq:certainty_equivalent} from Corollary~\ref{cor:certainty_equivalent}.

\section{Details Regarding the Proof of Lemma~\ref{lemma:completing_convergence}}
\label{sec:details_convergence_proof}
Rearranging Equation~\eqref{eq:upper_value_bound} and averaging over $M$ iterations, one obtains the following:
\begin{equation}
\label{eq:details_convergence}
\frac{1}{M} \sum_{m=0}^{M-1} \left( \mathbb{E}_{p} \left[ B_{\star} V(\bm{s}) \right] - \mathbb{E}_{p} \left[ B_{\pi^{(m)}_{\text{prior}},\pi^{(m+1)}} V(\bm{s}) \right]  \right) \leq \frac{1}{M \beta} \mathbb{E}_{p,\pi^{\star}} \left[ \log \frac{\pi^{(M)}(\bm{a}|\bm{s})}{\pi^{(0)}(\bm{a}|\bm{s})} \right],
\end{equation}
which leads directly to Equation~\eqref{eq:completing_convergence} by taking the max over actions and using $\pi^{(M)}(\bm{a}|\bm{s}) \leq 1$.

\section{Remark Regarding Lemma~\ref{lemma:completing_convergence}}
\label{sec:remark_completing_convergence}

Lemma~\ref{lemma:completing_convergence} also suggests to initialize $\pi^{(0)}$ with a uniform distribution $\mathcal{U}(\mathcal{A})$ since it minimizes the upper bound in Equation~(\ref{eq:completing_convergence}), more formally $\mathcal{U}(\mathcal{A}) = \argmin_{\pi} \max_{\bm{a}} \log \left( \pi(\bm{a})\right)^{-1}$ for all admissible probability distributions $\pi$ over the action set $\mathcal{A}$.

\section{Grid World Setup}
\label{sec:grid_world}

In the grid world example from Figure~\ref{fig:grid} in Section~\ref{sec:theorie} from the main paper, the agent has to reach a goal in the bottom left of a $16 \times 16$ grid. Reaching the goal is rewarded with $+9$ and terminates the episode, whereas each step is penalized with $-1$. The agent can take five actions in each state, i.e. $\mathcal{A} = \{\text{'left'}, \text{'right'}, \text{'up'}, \text{'down'}, \text{'stay'} \}$. The environment is deterministic and the discount factor is $\gamma = 0.9$. The stopping criterion for the value iteration scheme is when the infinity norm of the difference value vector of two consecutive iterations drops below $5 \cdot 10^{-3}$ (for both soft value iteration as well as mutual-information-regularized value iteration). The stopping criterion for the inner Blahut-Arimoto scheme that is necessary in order to apply $B_\star$ (required for one value iteration step) is when the maximum absolute difference in probability values in two consecutive inner iterations drops below $5 \cdot 10^{-3}$.

\section{Experiment Details for Mujoco}
\label{sec:implementation_details}

All function approximators are trained with the Adam optimizer~\cite{Kingma2015} using a learning rate of $3\cdot10^{-4}$. The discount factor is $\gamma=0.99$ and the replay buffer can store at most one million transition tuples. The minibatch size for training is $256$ for all objectives. All deep networks have two hidden layers with $256$ $\relu$-activations implemented with PyTorch. When updating Q-values, we use an exponentially averaged V-target network with time parameter $\tau=0.01$~\cite{Haarnoja2018}. The reward scale $\beta$ is set to $10$ for all experiments (also for the SAC baseline to ensure a fair comparison). Marginal policy values are approximated with $20$ samples from the uniform Gaussian. We also use a separate replay buffer for training the marginal policy with buffer sizes depending on the environment (Pendulum: $10^3$; InvertedDoublePendulum, Swimmer, Reacher: $10^4$; Hopper, Walker2d, Ant, Humanoid: $5\cdot10^4$). Hyperparameters for the baselines DDPG, PPO and SAC are close to the literature~\cite{Schulman2017b,Fujimoto2018,Haarnoja2018} but with the same neural network architectures as above for a fair comparison.

The state value network $V_{\psi}$ and the state-action value network $Q_{\theta}$ are ordinary feedforward networks that output scalar values. The policy network $\pi_{\phi}$ and the marginal policy network $\pi_{\chi}$ output a mean and a log standard deviation vector encoding a Gaussian. Since actions in Mujoco are bounded, unbounded actions are adjusted through a $\tanh$-nonlinearity accordingly.

We also experimented with a proper variational autoencoder~\cite{Kingma2014} to learn a generative model for marginal actions with an ELBO objective. The autoencoder requires an additional parametric recognition model---in our case a two-hidden-layer neural network (like the ones mentioned earlier) that maps actions to a mean and a log standard deviation vector representing a Gaussian in latent space. 

\section{Experiment Results in Mujoco}
\label{sec:raw_results}

The kink in performance of MIRACLE in Ant is due to the fact that in one out of $10$ runs, episodic rewards significantly started dropping after around $800,000$ to $900,000$ steps, while in all the other runs, the policies kept improving. Also note that there is no DDPG baseline in Ant. In initial experiments on Ant, we observed that for our hyperparameter setting, the RLlib implementation immediately dropped to large negative reward values and never recovered from there.

\begin{figure}[ht]
\begin{center}
\centerline{\includegraphics[width=1.0\columnwidth]{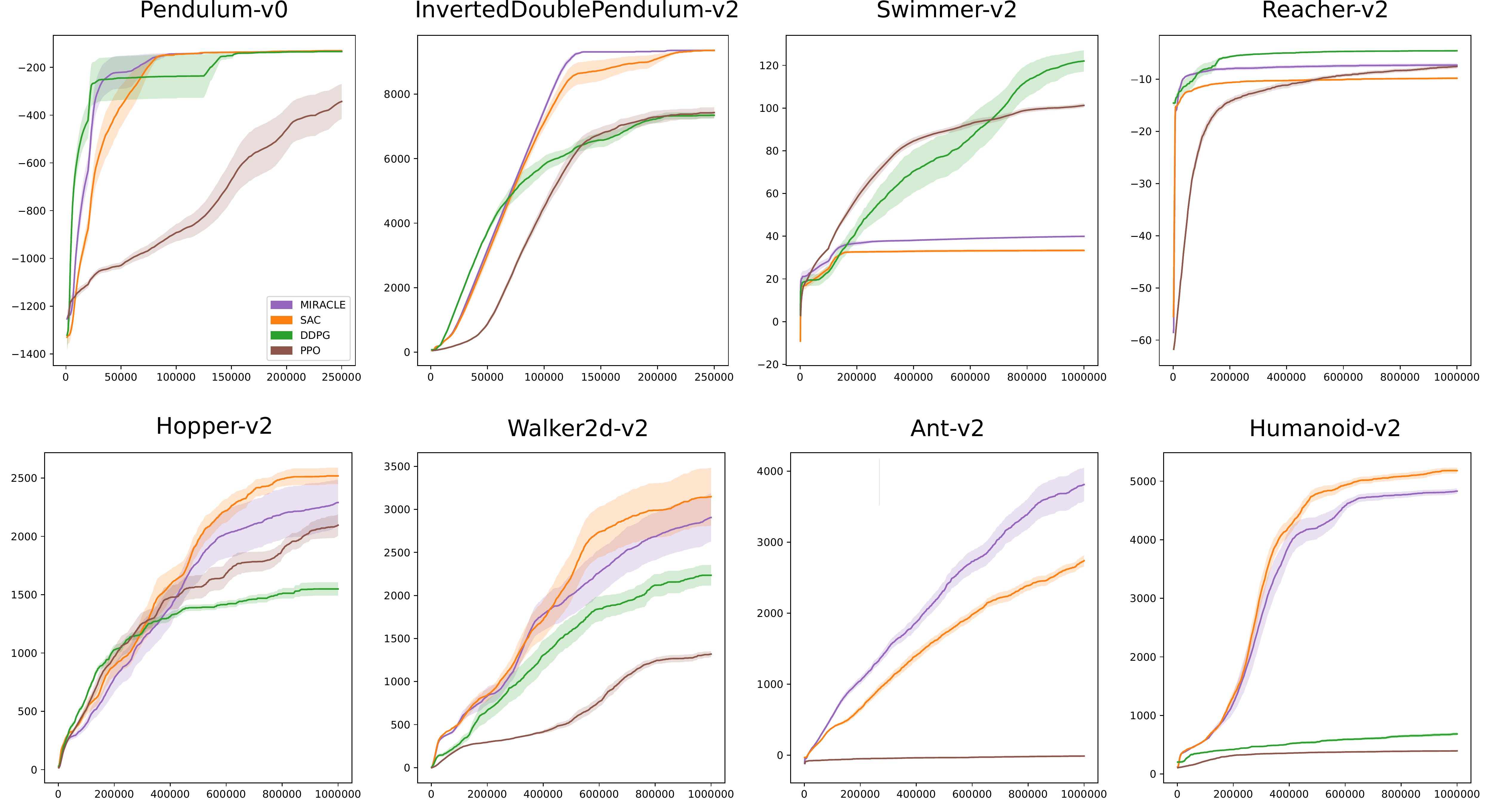}}
\caption{Mujoco Experiments Best Rewards. The plot shows the same experiments as in Figure~\ref{fig:best_reward} from the main paper, but reports the best episodic reward obtained so far during training (averaged over the last $100$ episodes). Under this metric, MIRACLE clearly improves over SAC in five out of eight environments, most notably Ant.}
\label{fig:reward}
\end{center}
\end{figure}

\begin{figure}[ht]
\begin{center}
\centerline{\includegraphics[width=1.0\columnwidth]{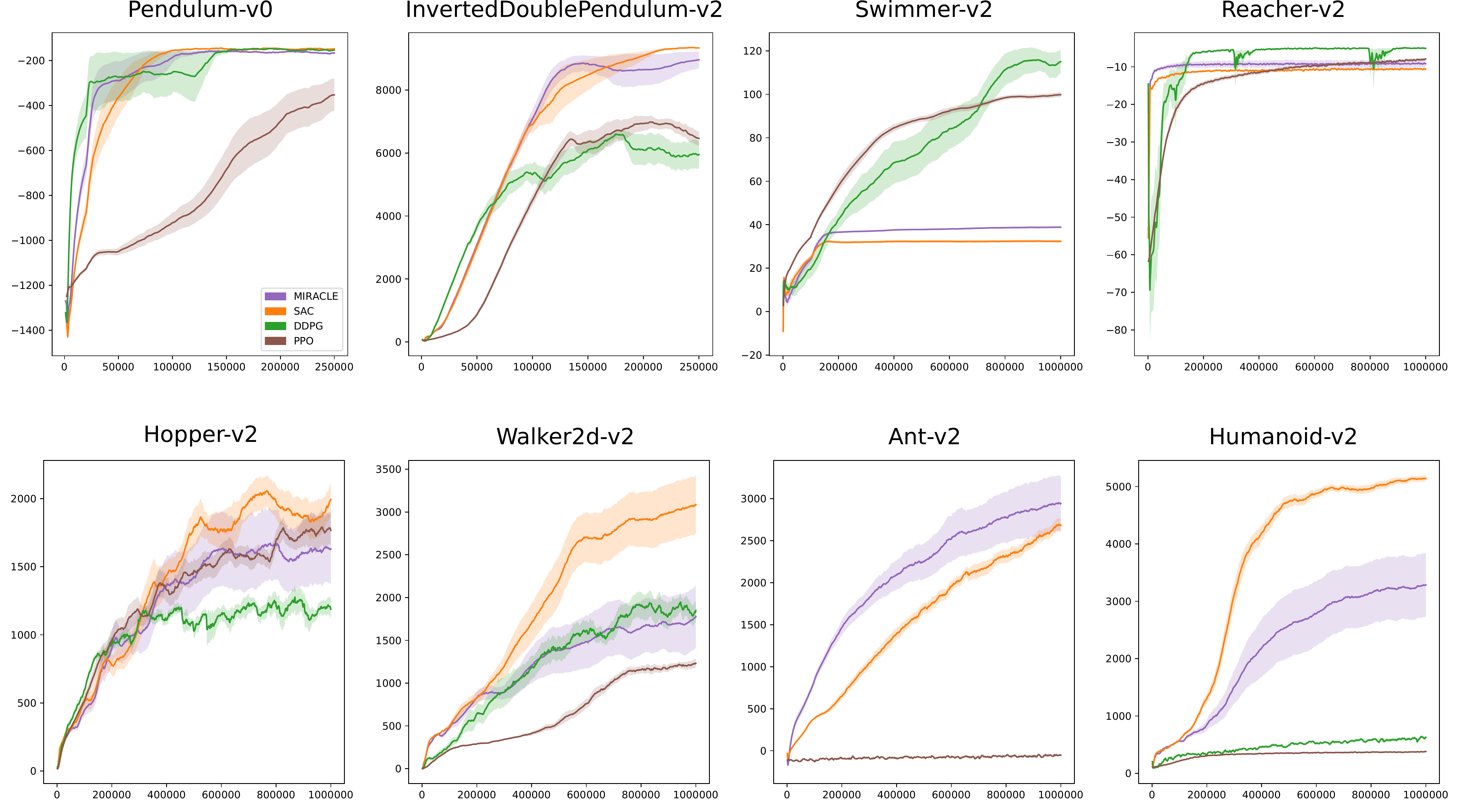}}
\caption{Mujoco Variational-Inference Experiments. The plot reports experiments with a variational autoencoder to model marginal actions. Results are depicted the same way as in Figure~\ref{fig:best_reward} from the main paper. Overall, results are worse compared to the main paper. The experimental setup was however exactly the same, i.e. both the generative and the recognition model performed a single parameter update for one minibatch only in each step. We hypothesize a better result could be obtained under a different training scheme that allows the autoencoder more training samples.}
\label{fig:reward_vi}
\end{center}
\end{figure}

\begin{figure}[ht]
\begin{center}
\centerline{\includegraphics[width=1.0\columnwidth]{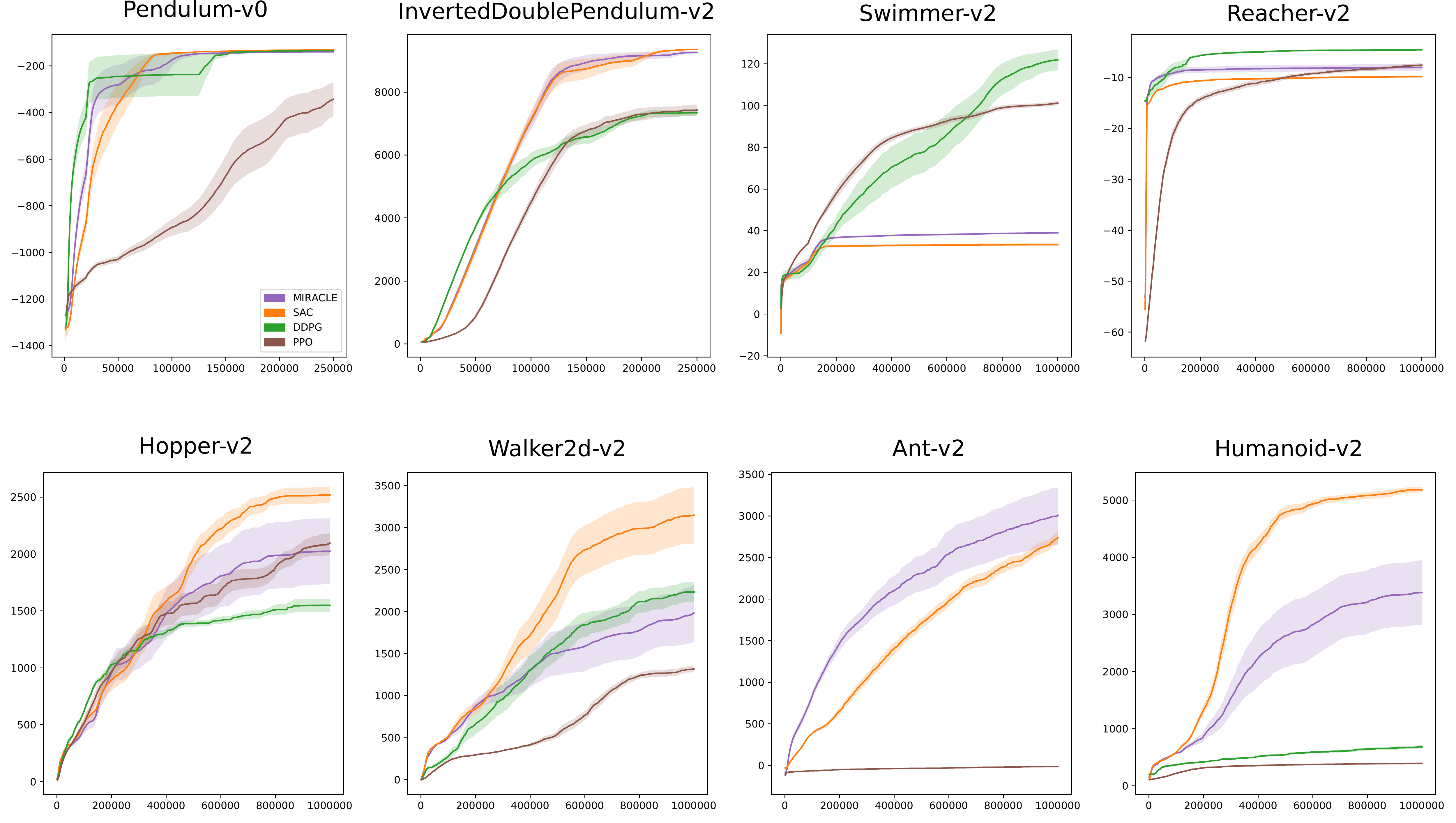}}
\caption{Mujoco Variational-Inference Experiments Best Rewards. Results are depicted similar to Figure~\ref{fig:reward} but for the variational autoencoder experiments.}
\label{fig:reward_cumul_vi}
\end{center}
\end{figure}

Here, we also report in Figure~\ref{fig:reward} the maximum value obtained so far (in the course of training) of the average episodic reward over the last $100$ episodes~\cite{Chua2018}, as opposed to the main paper that reports the raw version of this metric. Reporting results with this metric highlights improvements of MIRACLE over baselines more clearly. Additional results similar to Figures~\ref{fig:best_reward} and~\ref{fig:reward} for experiments with variational autoencoders to model marginal actions are shown in Figures~\ref{fig:reward_vi} and~\ref{fig:reward_cumul_vi}. These were conducted under the same setup as the other experiments and yielded worse results. Better results might be obtained by a different training procedure, e.g. more training samples for the autoencoder.

\end{document}